\documentclass{article}
\usepackage{iclr2027_conference,times}
\usepackage{amsmath,amsfonts,bm}

\def\eqref#1{equation~\ref{#1}}

\def\1{\bm{1}}

\DeclareMathAlphabet{\mathsfit}{\encodingdefault}{\sfdefault}{m}{sl}
\SetMathAlphabet{\mathsfit}{bold}{\encodingdefault}{\sfdefault}{bx}{n}

\usepackage{graphicx}
\usepackage{booktabs}
\usepackage{capt-of}
\usepackage{hyperref}
\usepackage{url}

\newcommand{\methodname}{ForcingBooster}
\graphicspath{{fig/}}
\iclrfinalcopy

\title{\methodname{}: Adaptive Patch-wise Denoising
for Fast Autoregressive Video Diffusion}
\author{
    Zekai Gu\textsuperscript{1,2} \quad
    Peng Wang\textsuperscript{2}\thanks{Project leader.} \quad
    Jiahao Lu\textsuperscript{1} \quad
    \textbf{Yongsen Mao}\textsuperscript{1} \quad
    \textbf{Cheng Lin}\textsuperscript{3} \quad
    \textbf{Yuan Liu}\textsuperscript{1}\thanks{Corresponding author.}
    \\[0.5em]
    \normalfont
    \textsuperscript{1}HKUST \quad
    \textsuperscript{2}VAST  \quad
    \normalfont
    \textsuperscript{3}Macau University of Science and Technology
}

\begin{document}

\maketitle
\pagestyle{plain}
\thispagestyle{plain}

\begin{abstract}
Autoregressive video diffusion models have become a leading approach to real-time interactive video generation and video world modeling. However, existing methods typically allocate computation uniformly across space when generating the next video chunk, updating all patches with the same denoising schedule and number of function evaluations. This ignores the spatial heterogeneity of video content: static, simple, or recurring regions are often easier to denoise, while other regions require further refinement or additional context. A uniform budget thus leads to redundant computation on easy patches, whereas reducing the fixed step count can leave difficult patches insufficiently refined. We introduce \methodname{}, a general framework that dynamically allocates denoising steps to individual patches: easy patches stop early to improve efficiency, while difficult patches receive more refinement steps. This design accelerates generation with negligible quality degradation. Our method is compatible with most existing few-step autoregressive diffusion models, requiring only a lightweight routing module with modest training. Experiments demonstrate that \methodname{} reduces the total number of denoising steps applied across video patches by about 50\% while maintaining visual quality comparable to the four-step baseline. Project page: \href{https://igl-hkust.github.io/ForcingBooster}{https://igl-hkust.github.io/ForcingBooster}
\end{abstract}

\begin{figure}[!ht]
    \centering
    \includegraphics[width=\linewidth]{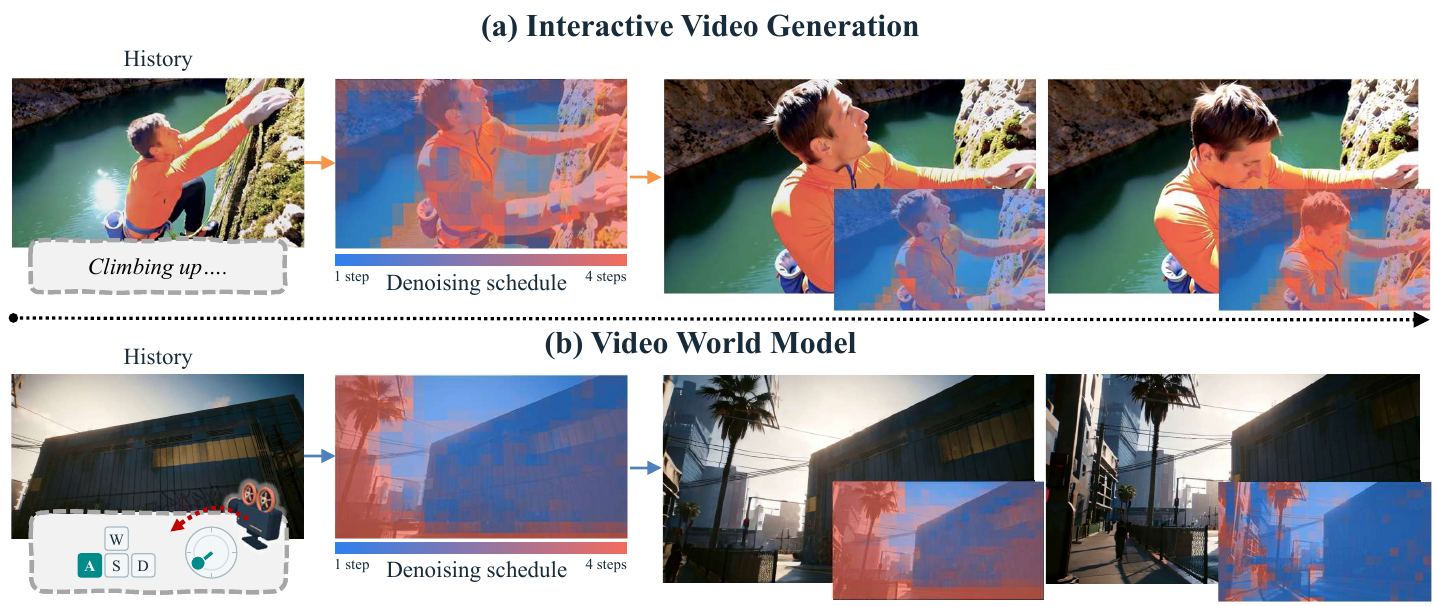}
    \caption{\methodname{} accelerates interactive video generation and world model by adaptively allocating different denoising steps to different patches. Heatmap overlays visualize router allocations from matching conditional samples: \textbf{static backgrounds need fewer updates than moving objects, new regions or detailed textures.}}
    \label{fig:teaser}
\end{figure}

\section{Introduction}
\label{sec:introduction}

Video generation~\cite{wan2025wanopenadvancedlargescale,hacohen2026ltx2efficientjointaudiovisual} is becoming a core approach for digital content creation~\cite{liang2025diffusionrenderer}, scene simulation~\cite{sun2024drivescenegen,wen2024panacea}, and interactive video world modeling~\cite{robbyant2026lingbotworld,yuan2026fastwam}. Despite their different goals, these applications all require models to preserve visual quality while responding to input signals and generating future content with low latency. Recent video diffusion models have substantially improved generation quality, yet their iterative sampling process remains too slow for real-time interaction. Bridging this quality--latency gap is therefore a central challenge for real-time video generation and video world models.

Autoregressive video diffusion converts joint full-video generation into a sequence of next-chunk predictions by introducing causal attention. These methods~\citep{chen2024diffusionforcing,guo2025selfresampling} partition a video into short temporal chunks and generate each chunk conditioned on the preceding ones. Once the first chunk is generated, the model can begin producing output and continue with subsequent chunks. This allows the model to output completed chunks without waiting for the entire video. Causal attention enables the use of KV caching to avoid repeatedly processing previous chunks. Distillation methods~\cite{yin2025causvid,huang2025selfforcing,zhu2026causalforcing,zhao2026causalforcingplusplus} further reduce the generation time of each chunk, compressing models that require dozens of denoising steps into autoregressive generators that use four steps or fewer. Recent methods reduce computation by compressing historical context~\cite{yuan2026helios} or choosing the number of denoising steps for entire frames or chunks~\cite{le2026dsa}. 

Although these methods substantially reduce the denoising iterations required for each frame or chunk, they typically assign the same denoising schedule and computational budget to all latent patches within that chunk. Every new chunk still requires a complete multi-step diffusion process. Some work has explored different denoising schedules across spatial regions in reasoning and image generation~\cite{wewer2025spatialreasoning,schusterbauer2026patchforcing}. Our analysis indicates that some patches produce predictions close to the full denoising result at earlier steps, while others still require further refinement. For example, as shown in Fig.~\ref{fig:teaser}(a), fewer denoising steps are allocated to the background water (blue), while more are allocated to the moving climber and the textured rock face (red). In Fig.~\ref{fig:teaser}(b), as the camera pans left, the sky and the large building on the right receive fewer steps, whereas the trees and newly revealed structures on the left receive more. Applying all steps to every patch therefore wastes computation on patches that are already stable, whereas reducing the step count for the entire chunk can degrade regions that still need refinement. Consequently, even when the base model has already been distilled to very few steps, allocating computation according to local content still offers substantial room for improvement.

We propose \methodname{}, which equips an existing few-step autoregressive generator with a lightweight router that dynamically determines which latent patches require further denoising at each step. The router examines each patch in the context of the generated history and the current conditioning signals, such as camera poses or text controls, to assess whether its prediction has already sufficiently converged. Patches deemed complete stop denoising, while only the remaining uncertain or condition-sensitive regions continue to be refined. In this way, \methodname{} allocates computation adaptively across space and denoising steps, reducing redundant refinement without modifying the generator itself.

Two key challenges arise in enabling patch-wise early stopping. First, if a patch terminates before completing the full denoising schedule, we need a valid clean latent that can be directly used as its output. To address this, we observe that diffusion and flow-matching generators admit a clean-latent estimate at every intermediate denoising step, either predicted directly or recovered from their native parameterization. This intermediate $\hat{\mathbf{x}}_0$ prediction provides a natural output for early-exited patches, allowing different regions to terminate at different denoising steps without modifying the original sampling schedule.

Second, there is no ground-truth supervision indicating when each patch should stop. We therefore formulate router learning as a trade-off between generation quality and computation. Specifically, we introduce a stopping surrogate loss that penalizes high stop probabilities when the current clean-latent prediction deviates from the full-step denoised latent, together with a computational loss that encourages patches to terminate with fewer denoising steps. These objectives are optimized with on-policy rollouts of the current router, so the estimated continuation cost reflects its evolving routing behavior. The full-step outputs of the frozen generator serve as training references, enabling the router to be trained without additional video data or manually defined stopping labels.

We validate \methodname{} across video generation and world models with different scales and architectures, ranging from camera-conditioned world models to real-time interactive autoregressive video generators. On average, the router requires only approximately 500 training steps and reduces the total number of denoising steps applied across video patches by about 50\%, while maintaining VBench scores comparable to those of the corresponding four-step baselines.

\section{Related Work}
\label{sec:related-work}

\subsection{Autoregressive Video Diffusion}

Autoregressive video diffusion generates temporal chunks conditioned on earlier ones. Diffusion Forcing trains causal models with independent noise levels across tokens~\cite{chen2024diffusionforcing}. Later work improves long video generation through history management and noise scheduling~\cite{yang2025longlive,liu2025rollingforcing}, or trains on generated histories to reduce error accumulation~\cite{po2025bagger,guo2025selfresampling,cui2025selfforcingplusplus}. Other work studies multiplayer world modeling~\cite{savva2026solaris} and flexible chunking and generation orders~\cite{ma2026flexforcing}.

\subsection{Diffusion Distillation and Post-Training}

Few-step distillation reduces the number of denoiser evaluations. DMD and DMD2 distill diffusion models through distribution matching~\cite{yin2024dmd,yin2024dmd2,zhao2026perceptualflowmatchingfewstep}, and CausVid extends this approach to autoregressive video generation~\cite{yin2025causvid}. Self Forcing trains on generated histories~\cite{huang2025selfforcing}, while Causal Forcing uses an autoregressive teacher for initialization~\cite{zhu2026causalforcing}. Further work studies generation in one or two steps~\cite{zhao2026causalforcingplusplus,feng2026oneforcing}, flexible sampling budgets~\cite{gu2026anyflow}, diversity preservation~\cite{chen2026dataforcing}, and reward supervision~\cite{lu2025rewardforcing}. Recent on-policy self-distillation (OPSD) methods use richer teacher context to supervise the model's own rollouts~\cite{jiang2026dopsdonpolicyselfdistillationcontinuously,liu2026opsdvonpolicyselfdistillationposttraining}, while DanceOPD~\cite{zhou2026danceopdonpolicygenerativefield} learns from expert velocity fields on student-generated states. These methods update the generator and typically use a shared step budget across spatial patches within each frame or chunk.

\subsection{Adaptive Computation for Diffusion}

Adaptive sampling selects denoising budgets for individual samples~\cite{zhang2025adadiff}, while spatial reasoning models allow different regions to follow different denoising schedules~\cite{wewer2025spatialreasoning}. Other methods reduce computation within each denoiser evaluation~\cite{zhao2025dydit,you2025diffcr} or process fewer tokens through history compression and multiscale generation~\cite{yuan2026helios}.

Caching methods reuse intermediate features across denoising steps~\cite{ma2024deepcache,lyu2025fastercache}. TeaCache estimates output changes from inputs modulated by timestep embeddings to guide reuse~\cite{liu2025teacache}, while AdaCache adapts cache intervals using video features~\cite{kahathapitiya2025adacache}. Other methods selectively recompute tokens or regions~\cite{zou2025toca,liu2026ras,xu2026motioncache}, extend caching to autoregressive video chunks~\cite{ma2026flowcache}, or adjust cache thresholds during denoising~\cite{agrawal2026acid}.

DSA~\cite{le2026dsa} jointly trains a confidence head and the generator to select denoising steps for entire frames or chunks, but spatial patches within each unit still share the same budget. Patch Forcing~\cite{schusterbauer2026patchforcing} trains image generators with spatially varying noise levels and uses predicted patch difficulty to guide sampling. HSA~\cite{chu2026hsa} assigns token-level step budgets without training and preserves KV context, while continuing to update skipped latents with cached velocities. In contrast, \methodname{} learns patch-wise stopping in a frozen few-step autoregressive generator, allowing easy regions to finish early while retaining further refinement for difficult ones. We train only a lightweight router and retain the original denoising schedule. Training balances reconstruction error against denoising cost using full-step references and histories generated under the current routing policy. Rather than advancing skipped latents with cached velocities, we fix each stopped patch's clean latent prediction. Sparse inference then removes it from subsequent Transformer updates once its KV cache is built. This reduces redundant computation without discarding the context needed by active patches and subsequent chunks.
\section{Method}
\label{sec:method}

\subsection{Preliminary and Overview}
\label{sec:method-overview}

\begin{figure}[t]
\centering
\includegraphics[width=\linewidth]{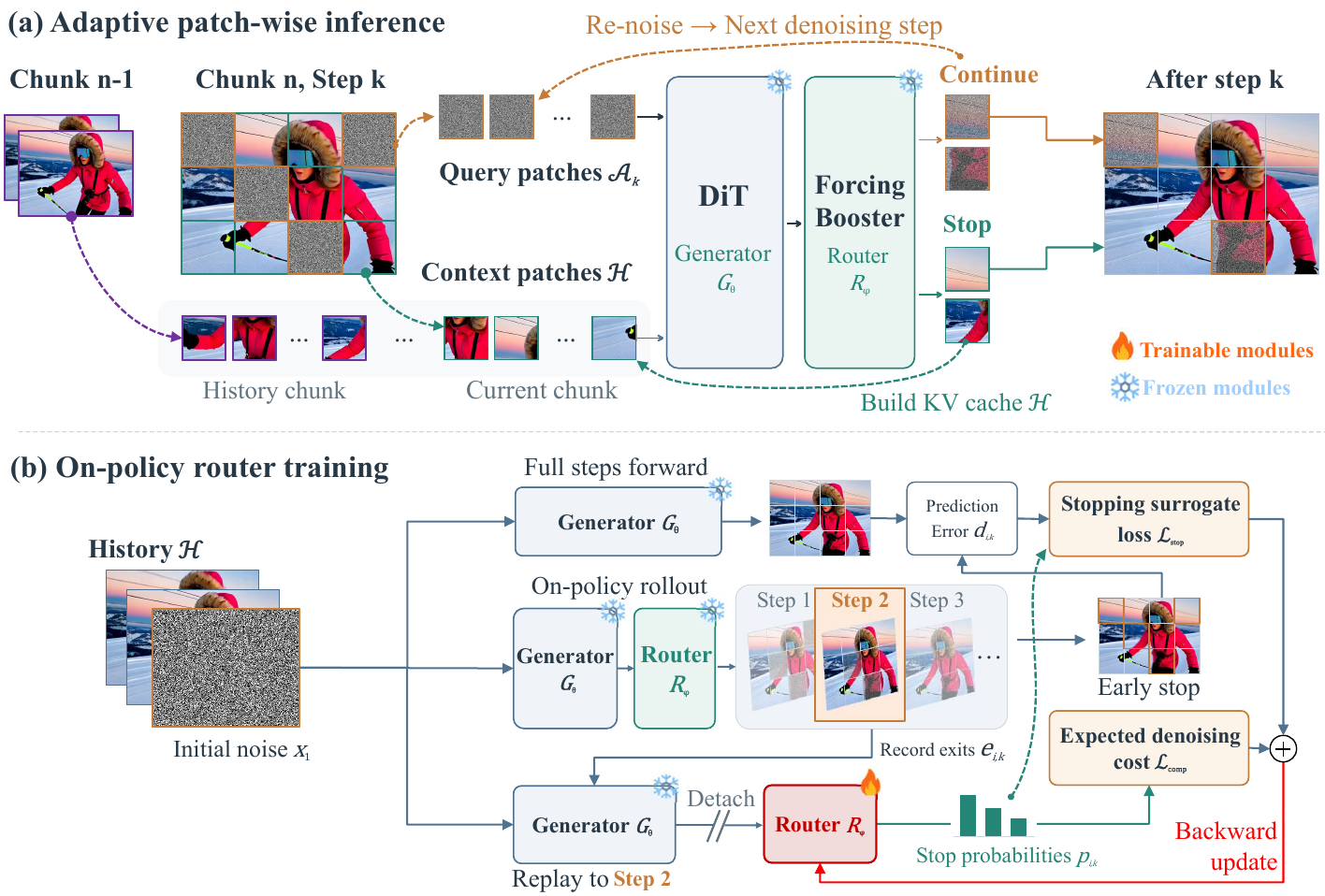}
\caption{\textbf{Adaptive patch-wise inference and on-policy router training.} (a) The router selects patches to continue denoising; stopped predictions provide reusable KV context. (b) A no-gradient rollout records exits for replay. Reconstruction and computational costs train the router through its stopping probabilities.}
\label{fig:pipeline}
\end{figure}

\methodname{} is designed to improve the efficiency of an existing autoregressive video generator with a router that allocates an adaptive denoising budget for different regions. Specifically, we will first introduce the $\mathbf{x}_0$-prediction mechanism of diffusion or flow matching models, which will be used in our denoising router to early terminate the denoising process for well generated patches.

\paragraph{$\mathbf{x}_0$-prediction of autoregressive video generator.} The generator produces videos chunk by chunk in latent space in an autoregressive manner. Each chunk starts from Gaussian noise $x_1$ and follows a fixed schedule $\{t_k\}_{k=1}^{K}$ with $t_0=1,\ t_K=0$. We express the denoising process at step $k$ as
\begin{equation}
\mathbf{\hat{x}}_{0,t_{k+1}}
=G_{\theta}\left(\mathbf{x}_{t_k},t_k;\mathcal{H},\mathbf{y}\right),
\qquad 
\mathbf{x}_{t_{k+1}}=\mathcal{E}_{t_{k+1}}(\mathbf{\hat{x}}_{0,t_{k+1}})
\qquad k=1,\ldots,K,
\label{eq:denoising}
\end{equation}
where $G_{\theta}$ is the denoising network with parameters $\theta$, $\mathbf{x}_{t_k}$ is the noisy input at noise level $t_k$, $\mathbf{y}$ denotes generation conditions, $\mathcal{H}$ denotes the generation history represented by the KV cache of previously generated frames, $\mathbf{\hat{x}}_{0,t_{k+1}}$ is the predicted clean latent at this step, and $\mathcal{E}_{t_{k+1}}$ maps this clean prediction to the next noisy state $\mathbf{x}_{t_{k+1}}$. This formulation abstracts over the specific parameterization used by diffusion or flow-matching models: even when the network predicts noise or velocity, an equivalent clean-latent estimate $\mathbf{\hat{x}}_{0,t_{k+1}}$ can be obtained at each step. Importantly, this intermediate clean prediction allows a patch to terminate denoising early and directly use its current $\hat{\mathbf{x}}_{0,t_k}$ as the final latent prediction.

\paragraph{Denoising router.} 
A vanilla autoregressive video generator will use the same number of denoising steps $K$ for every patch in the current generation chunk. However, different patches may require substantially different amounts of denoising budget. In particular, regions that can be reliably inferred from the generation history $\mathcal{H}$ may converge much earlier than newly revealed or condition-dependent regions. To exploit this redundancy, \methodname{} applies a denoising router that predicts a stop probability $p_{i,k}$ for all patches after every denoising step. Then, for patches with large stop probability, we will stop denoising these patches in the subsequent steps, use their predicted clean latents $\mathbf{\hat{x}}_{0,t_{k}}$ at this step as the output, and further store the KV caches of these patches to $\mathcal{H}$ for the denoising of other non-stopped patches.

In the following, we describe how our router changes and speeds up the generation process (Section~\ref{sec:adaptive-patch-denoising}) and how we train our router with a surrogate loss (Section~\ref{sec:training-objective}).

\subsection{Denoising Steps Required by Different Patches}
\label{sec:patch-denoising-difference}

\begin{figure}[t]
\centering
\includegraphics[width=\linewidth]{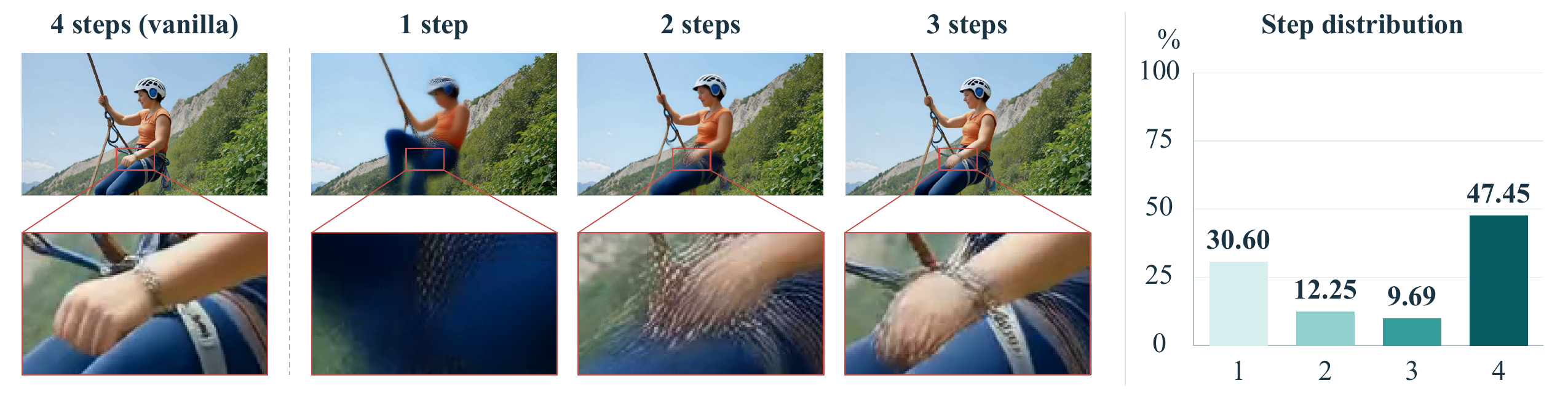}
\caption{\textbf{Denoising steps required by different patches.} Left: fixed-step generation results with enlarged local details. Right: patch-wise first-hit step distribution at $35\,\mathrm{dB}$ PSNR relative to the four-step reference in a separate controlled experiment.}
\label{fig:patch-denoising-difference}
\end{figure}

A few-step schedule reduces the number of denoising updates, but still applies the same step count to every patch. Some patches may already be close to the full-step output while others still need further updates. If this difference exists, we may be able to reduce computation selectively rather than shorten the schedule for all patches. We therefore conduct a controlled experiment to examine when each patch's prediction first approaches the full-step output.

We use Causal Forcing~\cite{zhu2026causalforcing} as the base model, fix the context to the first ground-truth video chunk, and generate only the immediately following target chunk. Each target chunk contains three latent frames. We hold the generation conditions, initial noise, and noise between denoising steps fixed and record the clean latent prediction at each step of a single four-step sampling run.

We decode each prediction and compute the peak signal-to-noise ratio (PSNR) against the decoded four-step output over the region corresponding to each patch. We record the first step at which this score reaches $35\,\mathrm{dB}$. This measures agreement with the generator's own full-step output, rather than reconstruction of the ground-truth video.

As shown in Fig.~\ref{fig:patch-denoising-difference} (right), $30.60\%$ of patches reach the threshold after the first step. A further $12.25\%$ and $9.69\%$ first reach it at the second and third steps, respectively. The remaining $47.45\%$ do not reach it within the first three steps.

Figure~\ref{fig:patch-denoising-difference} (left) shows the final frames from separate fixed-step generation runs with the same seed. Each run uses four steps for the first chunk and the indicated step count for subsequent chunks. The sky and hillside look broadly similar across the results, whereas the enlarged hand and forearm show clear differences. With one step, the hand is poorly defined. Two and three steps recover more of its shape, but visible distortions remain around the hand and wrist compared with the four-step result.

These results show that patches within the same chunk approach the full-step output at different rates. They motivate reducing later updates at positions already close to that output while allowing other positions to continue denoising.

\subsection{Adaptive Patch-wise Denoising Router}
\label{sec:adaptive-patch-denoising}

After each denoising step, the router predicts a stop probability for each patch to stop the denoising of well-generated patches. Then, we add the KV cache of these stopped patches to the history $\mathcal{H}$ and proceed to the next denoising step.

\paragraph{Stop probability prediction.} Let $\mathcal{A}_k$ denote the active patches at the beginning of step $k$ and $\mathcal{A}_{0}$ includes all patches. After the denoising step $k$, the router $R_{\phi}$ reuses hidden features $\mathbf{u}_k$ from the generator's last transformer layer to predict the stop probability $p_{i,k}$ by
\begin{equation}
p_{i,k}=\left[R_{\phi}(\mathbf{u}_k)\right]_i,
\qquad i\in\mathcal{A}_k.
\end{equation}
where $\phi$ denotes the router parameters. A threshold $\eta$ is used to determine the active patches $\mathcal{A}_{k+1}$
\begin{equation}
\mathcal{A}_{k+1}
=\left\{i\in\mathcal{A}_k\mid p_{i,k}\leq\eta\right\},
\qquad k<K.
\end{equation}

As shown in Fig.~\ref{fig:pipeline}(a), only $\mathcal{A}_{k+1}$ will be used in the denoising of the next step. For patches stopped at this step, their $\mathbf{\hat{x}}_{0,t_{k+1}}$ from Eq.~(\ref{eq:denoising}) are used as output. Meanwhile, the KV values of these stopped patches will be stored into the KV cache $\mathcal{H}$ to help the denoising of the remaining patches.

\paragraph{Computation reduction.}
The router progressively reduces the number of active patches throughout denoising. Once a patch stops, it is removed from subsequent denoising updates and its key-value representations are cached for reuse. Therefore, later denoising steps only compute new representations for the remaining active patches, while stopped patches continue to provide contextual information through the KV cache without being recomputed. As more patches terminate, the active set shrinks monotonically, reducing the per-step computation and accelerating the generation.

\subsection{On-policy Router Training}
\label{sec:training-objective}

\paragraph{Learning the stop probability.}
Training the router is non-trivial because there is no ground-truth annotation indicating at which denoising step each patch should stop. Instead of constructing discrete stopping labels, we directly relate the stop probability to the quality of the current clean-latent prediction. Intuitively, if the predicted clean latent $\hat{\mathbf{x}}_{0,t_k}$ at step $k$ is already close to the target clean latent, further denoising brings little benefit and the router should assign a high stop probability. Conversely, a large prediction error indicates that the patch still requires further refinement and should receive a low stop probability. Based on this observation, we introduce a differentiable surrogate loss and a denoising cost loss to train the router.

\paragraph{Stopping surrogate loss.}
Let $\mathbf{x}_0^{*}$ denote the target clean latent, and let
\begin{equation}
d_{i,k}
=
\left\|
\hat{\mathbf{x}}_{0,t_k}^{\,i}
-
\mathbf{x}_0^{*,i}
\right\|_2^2
\end{equation}
measure the prediction error of patch $i$ after step $k$. We penalize stopping in proportion to this error:
\begin{equation}
\mathcal{L}_{\mathrm{stop}}
=
\frac{1}{|\mathcal{A}_k|}
\sum_{i\in\mathcal{A}_k}
p_{i,k}\, d_{i,k}.
\end{equation}
This objective discourages a high stop probability when the current prediction is still inaccurate, while imposing little penalty on stopping patches with converged clean-latent predictions. 

\paragraph{Expected denoising cost.}
Optimizing the surrogate loss alone will lead to a trivial solution in which the router assigns a zero stop probability to every patch, allowing all patches to run through the complete denoising schedule. To encourage early stopping, we introduce a denoising cost loss that minimizes the expected number of denoising steps for each patch. For a candidate patch $i$ at step $k$, stopping immediately incurs a cost of $k$, whereas continuing denoising incurs a cost determined by the final stopping step $e_{i,k}$ of our routing policy. The denoising cost is then
\begin{equation}
\mathcal{L}_{\mathrm{comp}}
=
\frac{1}{|\mathcal{A}_k|}
\sum_{i\in\mathcal{A}_k}
\left[
p_{i,k} \cdot k
+
(1-p_{i,k}) \cdot e_{i,k}
\right].
\end{equation}
Here, $p_{i,k}\cdot k$ corresponds to the cost of stopping at the current step, while $(1-p_{i,k})\cdot e_{i,k}$ represents the cost of continuing until the patch eventually exits. Since $e_{i,k}>k$, minimizing this objective favors a larger stop probability whenever further denoising is unnecessary, thereby reducing the overall denoising computation. To estimate $e_{i,k}$, we roll out the remaining denoising steps using the current router and stopping threshold $\eta$, and record when each active patch eventually exits. 
This on-policy estimation allows the computational cost to reflect the behavior of the current routing policy as it evolves during training.

\paragraph{Self-generated training targets.}
The target $\mathbf{x}_0^{*}$ can be obtained directly from the frozen generator itself, without requiring additional video data or stopping annotations. Specifically, we first run the original generator with its full denoising schedule and use the resulting clean latent as the training target, and record this trajectory. We then use this trajectory perturb this latent to a randomly sampled noise level $t_k$ and feed the noisy latent, which is recorded in the trajectory, back to the frozen generator to obtain the intermediate clean prediction $\hat{\mathbf{x}}_{0,t_k}$ and the corresponding hidden features for router prediction. We refer to this process as "replay." The discrepancy between $\hat{\mathbf{x}}_{0,t_k}$ and the full-step result therefore directly measures how much quality would be lost by stopping at step $k$. In this way, the pretrained generator provides both the training inputs and supervision for the router, allowing it to be trained without collecting any additional data.

\begin{figure}[t]
\centering
\includegraphics[width=\linewidth]{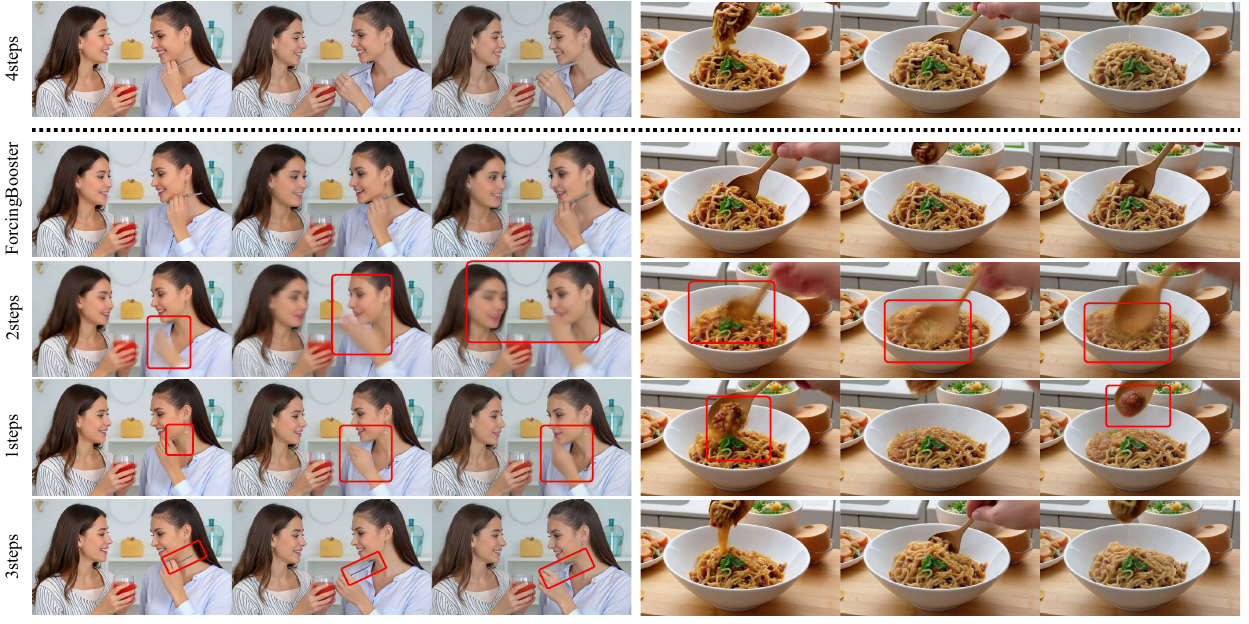}
\caption{\textbf{Qualitative comparison.} The rows show the results from four-step baseline to one-step sampling, and \methodname{}. One-step and Two-step sampling exhibits a decline in quality and visible artifacts. Red boxes indicate visual artifacts.}
\label{fig:exp1}
\vspace{-18pt}
\end{figure}
\section{Experiments}
\label{sec:experiments}

We evaluate \methodname{} on several video generation methods and world models, comparing generation quality and inference time with uniform fixed-step sampling. We then analyze spatial computation allocation and router settings.

\subsection{Implementation Details}
\label{sec:experimental-setup}

Our main experiments use Causal Forcing 1.3B~\citep{zhu2026causalforcing} to generate 81-frame videos at $480\times832$ and 16 FPS. Following its training data setup, we train only the router for 500 steps using AdamW, a learning rate of $10^{-4}$, and a global batch size of 8. The pretrained components and denoising schedule remain fixed. Following common practice~\cite{gu2026anyflow,huang2025selfforcing,zhu2026causalforcing}, we adopt VBench~\cite{huang2024vbench} to evaluate generated videos across 16 fine-grained dimensions. We report the aggregated Quality and Semantic scores, together with their weighted combination, the Total score. For this model, inference uses a single A800 GPU with BF16 precision and batch size 1. We report mean end-to-end generation time, including text encoding, the first chunk, router execution, KV cache construction, and VAE decoding.

\subsection{Main Results}
\label{sec:main-results}

We compare adaptive patch-wise denoising with uniform sampling using one to four steps. The original four-step sampler (vanilla) without routing serves as the baseline. All methods use four steps for the first chunk and apply their respective strategies to subsequent chunks. Fixed-step baselines execute a prefix of the original schedule and return the clean latent prediction from the last executed step. Generation conditions and noise are matched across methods.

\begin{table}[t]
\centering
\caption{Comparison on Causal Forcing checkpoint. Time includes generation through VAE decoding. Speedup and denoising patch-NFE reduction are relative to each
model's four-step baseline.}
\label{tab:cf13b-main-results}
\small
\setlength{\tabcolsep}{4pt}
\begin{tabular}{lrrrrrr}
\toprule
Method & Quality $\uparrow$ & Semantic $\uparrow$ & Total $\uparrow$ & Time (s) $\downarrow$ & Speedup $\uparrow$ & NFE Reduction $\uparrow$ \\
\midrule
1 step & 83.06 & 69.50 & 80.35 & 7.08 & $1.58\times$ & 64.29\% \\
2 steps & 83.69 & 71.66 & 81.29 & 8.67 & $1.29\times$ & 42.86\% \\
3 steps & 84.24 & 71.96 & 81.78 & 9.78 & $1.14\times$ & 21.43\% \\
4 steps (baseline) & 84.54 & 72.21 & 82.08 & 11.15 & - & - \\
\methodname{} & 83.88 & 72.20 & 81.55 & 8.56 & $1.30\times$ & 50.19\% \\
\bottomrule
\end{tabular}
\vspace{-10pt}
\end{table}

Table~\ref{tab:cf13b-main-results} shows that \methodname{} achieves a $1.3\times$ end-to-end speedup reduces the total number of denoising steps applied across video patches by about 50\%, while retaining Quality and Total scores close to the four-step baseline and an almost unchanged Semantic score. At a runtime similar to uniform two-step sampling, it also achieves higher scores on all three metrics, showing the benefit of allocating denoising steps across patches rather than reducing them uniformly. The qualitative comparison in Fig.~\ref{fig:exp1} reveals a clearer advantage of \methodname{} than the quantitative results suggest. The one-step, two-step, and three-step baselines all produce artifacts in regions with complex textures. In contrast, \methodname{} computes more on these patches and avoids artifacts.

\subsection{Further Evaluation}
\label{sec:more-methods}

We further evaluate Wan2.1 14B, LingBot-World-V2, and LongLive. Table~\ref{tab:cross-model-results} reports Quality, Semantic, Total, and speedup for each model's original four-step baseline, one-step, two-step baseline and \methodname{}. We compare methods within each base model to examine whether patch-wise computation allocation offers a useful trade-off across model scales and generation tasks.

\begin{table}[t]
\centering
\caption{Results on additional world model and interactive video generation methods. Speedups include VAE decoding and are relative to each model's four-step baseline. All methods use four steps for the first chunk.}
\label{tab:cross-model-results}
\small
\setlength{\tabcolsep}{4pt}
\begin{tabular}{lrrrr}
\toprule
Base model & Quality $\uparrow$ & Semantic $\uparrow$ & Total $\uparrow$ & Speedup $\uparrow$ \\
\midrule
Wan2.1 14B (1 step) & 82.10 & 71.35 & 79.95 & $1.96\times$ \\
Wan2.1 14B (2 step) & 84.71 & 77.93 & 83.35 & $1.29\times$ \\
Wan2.1 14B (4 steps) & 84.96 & 78.04 & 83.57 & - \\
Wan2.1 14B + \methodname{} & 84.73 & 77.65 & 83.32 & $1.31\times$ \\
\midrule
LingBot-World-V2 14B (1 step) & 78.79 & 69.72 & 76.98 & $1.98\times$ \\
LingBot-World-V2 14B (2 step) & 80.13 & 71.65 & 78.43 & $1.36\times$ \\
LingBot-World-V2 14B (4 steps) & 80.66 & 72.96 & 79.12 & - \\
LingBot-World-V2 14B + \methodname{} & 80.34 & 71.32 & 78.54 & $1.25\times$ \\
\midrule 
LongLive (1 step) & 78.91 & 73.18 & 77.76 & $1.54\times$ \\
LongLive (2 step) & 79.91 & 72.67 & 78.46 & $1.21\times$ \\
LongLive (4 steps) & 80.32 & 73.76 & 79.01 & - \\
LongLive + \methodname{} & 80.02 & 72.82 & 78.58 & $1.15\times$ \\
\bottomrule
\end{tabular}
\end{table}

\begin{figure}[t]
\centering
\includegraphics[width=\linewidth]{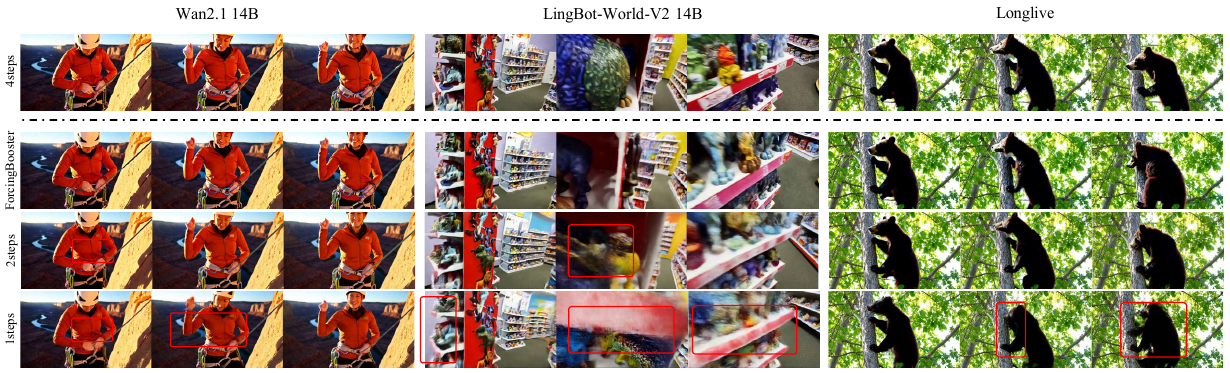}
\caption{\textbf{Qualitative comparison across additional models.} \methodname{} produces visual results comparable to the four-step baseline, while uniform one-step sampling often exhibits a decline in quality and visible artifacts. Red boxes indicate visual artifacts.}
\label{fig:exp2}
\end{figure}

Across these models, \methodname{} reduces runtime while maintaining Quality and Total scores close to the corresponding four-step baselines (Table~\ref{tab:cross-model-results}). It also achieves higher Quality and Total scores than uniform one-step sampling in every case. The qualitative comparison in Fig. ~\ref{fig:exp2} also indicates that one-step sampling results exhibit artifacts and a decline in quality across various models. Overall, adaptive patch-wise denoising achieves visual quality comparable to the four-step results while reducing inference time, and delivers superior visual quality and overall performance compared to uniform one-step sampling.

\subsection{Patch-wise Allocation}
\label{sec:ablations}

We compare different computation allocation strategies to examine whether the improvement in generation quality comes from effectively selecting spatial positions. Random preserves the distribution of exit steps within each latent frame but randomly permutes the spatial positions assigned to those steps. Chunk-wise router aggregates patch-wise exit decisions into a uniform number of denoising steps for each chunk. We compare these two strategies with \methodname{} and the original four-step baseline.

\begin{figure}[t]
\centering
\begin{minipage}[t]{0.49\textwidth}
\vspace{0pt}
\centering
\includegraphics[width=\linewidth]{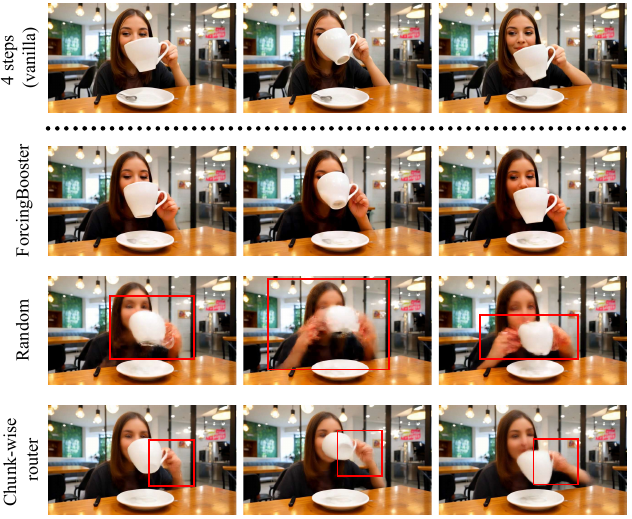}
\caption{\textbf{Qualitative comparison of computation allocation.} Columns show frames at matching time points under the same generation conditions. Red boxes indicate visual artifacts.}
\label{fig:patch-allocation-qualitative}
\end{minipage}\hfill
\begin{minipage}[t]{0.49\textwidth}
\vspace{0pt}
\centering
\setlength{\abovecaptionskip}{0pt}
\setlength{\belowcaptionskip}{6pt}
\captionof{table}{Comparison of computation allocation strategies. Time measures full-video generation, and speedups are relative to the four-step baseline.}
\label{tab:patch-allocation}
{\footnotesize
\setlength{\tabcolsep}{2.5pt}
\renewcommand{\arraystretch}{1.2}
\begin{tabular*}{\linewidth}{@{\extracolsep{\fill}}lrrr@{}}
\toprule
Method & Total $\uparrow$ & \shortstack{Time (s)$\downarrow$} & Speedup $\uparrow$ \\
\midrule
Random & 79.03 & 8.60 & $1.29\times$ \\
Chunk-wise router & 80.01 & 8.38 & $1.33\times$ \\
\methodname{} & 81.54 & 8.53 & $1.30\times$ \\
4 steps (baseline) & 82.07 & 11.13 & - \\
\bottomrule
\end{tabular*}
}
\end{minipage}
\vspace{-5pt}
\end{figure}

At comparable runtimes, \methodname{} achieves a higher Total score than both alternatives (Table~\ref{tab:patch-allocation}). The comparison with Random shows that the locations of early exits matter, not just their step-count distribution. The improvement over Chunk-wise router further supports assigning different step counts within the same chunk to preserve generation quality. Fig.~\ref{fig:patch-allocation-qualitative} shows the four methods at multiple time points under the same generation conditions. Chunk-wise router produces blurry frames in chunks assigned fewer denoising steps, with weaker details in regions such as the face and hands. Random exhibits severe flickering and detail loss in the generated video, with unstable local structures in the person and objects over time. In contrast, \methodname{} better preserves visual sharpness and temporal consistency, producing results closest overall to four-step baseline.

\subsection{Router Analysis}
\label{sec:quality-efficiency-ablation}

We study the computational cost weight $\lambda_{\mathrm{nfe}}$ and stopping threshold $\eta$. We measure agreement with the four-step output using LPIPS and report full-video runtime normalized by the four-step baseline.

\begin{figure*}[t]
\vspace{-4pt}
\centering
\includegraphics[width=0.95\linewidth]{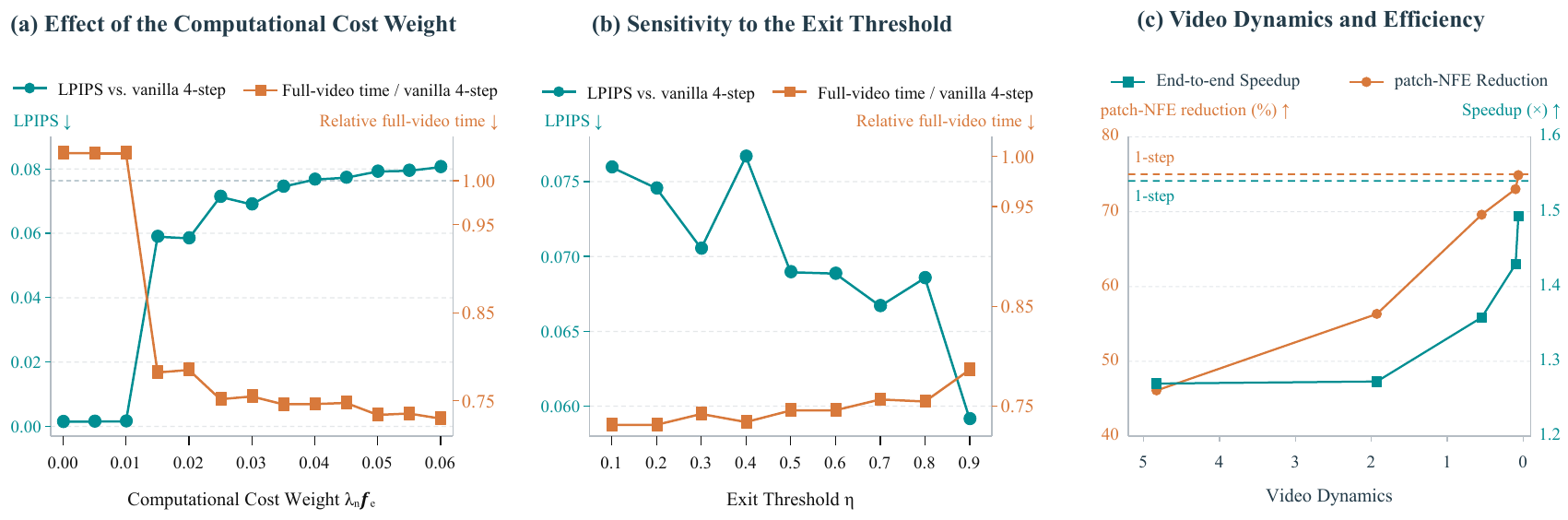}
\caption{\textbf{Router analysis.} \textbf{(a)} Effect of the computational cost weight $\lambda_{\mathrm{nfe}}$. \textbf{(b)} Effect of the stopping threshold $\eta$, with a separate router trained for each threshold and the same threshold used at inference. Both panels report LPIPS against the four-step output and normalized full-video runtime; lower is better.}
\label{fig:router_analysis}
\vspace{-18pt}
\end{figure*}

\paragraph{Computational cost.} We vary $\lambda_{\mathrm{nfe}}$ in the training objective from Section~\ref{sec:training-objective}. With zero or insufficient weight on $\mathcal{L}_{\mathrm{nfe}}$, the router retains all four steps and adds overhead without reducing runtime (Fig.~\ref{fig:router_analysis}(a)). Weights between $0.015$ and $0.030$ enable substantial time savings with low LPIPS. Further increases bring smaller runtime gains and larger deviations from the four-step output. These results show that a sufficient cost penalty is needed to learn effective early stopping, while overemphasizing it offers limited further acceleration.

\paragraph{Stopping threshold.} We vary $\eta$ from Section~\ref{sec:adaptive-patch-denoising}, training a separate router for each value from $0.1$ to $0.9$ with $\lambda_{\mathrm{nfe}}=0.030$. Each router uses the same threshold during training and inference. Across these routers, higher thresholds generally increase runtime but do not consistently lower LPIPS (Fig.~\ref{fig:router_analysis}(b)). Thus, a more conservative stopping threshold does not necessarily produce outputs closer to the four-step reference.

\paragraph{Video dynamics and speedup.}
We quantify video dynamics using RAFT~\cite{teed2020raftrecurrentallpairsfield} to estimate optical flow between adjacent frames, averaging its magnitude over space and time. We divide the videos into five equally sized groups by this measure and report the inference speedup and patch-wise NFE reduction for each group. As shown in Fig.~\ref{fig:router_analysis}(c), \methodname{} achieves larger speedups on lower-motion videos, approaching the runtime of one-step denoising with up to approximately $1.5\times$ speedup or reduces the total number of denoising steps applied across video patches by about 75\% over the four-step baseline.

\section{Conclusion}
\label{sec:conclusion}
We introduce \methodname{}, a general framework that dynamically allocates denoising steps to individual patches: simple patches terminate early to improve efficiency, whereas complex patches receive additional refinement steps. This design accelerates generation with negligible quality degradation. \methodname{} is compatible with most existing few-step autoregressive diffusion models and requires only a lightweight routing module and modest training, without fine-tuning the heavyweight generator or relying on additional video data.

\bibliography{references}
\bibliographystyle{iclr2027_conference}

\end{document}